\documentclass[10pt,twocolumn,letterpaper]{article}

\usepackage[pagenumbers]{cvpr} 

\newcommand{\model}{\textsc{Miburi}}

\definecolor{RDcolor}{rgb}{0.5, 0.1, 0.8}

\definecolor{bronze}{rgb}{1,1,0.6}
\definecolor{silver}{rgb}{0.969,0.796,0.600}
\definecolor{gold}{rgb}{0.941,0.592,0.600}
\newcommand{\gold}[1]{\colorbox{gold}{{#1}}}
\newcommand{\silver}[1]{\colorbox{silver}{{#1}}}

\definecolor{cvprblue}{rgb}{0.21,0.49,0.74}
\usepackage[pagebackref,breaklinks,colorlinks,allcolors=cvprblue]{hyperref}
\usepackage[accsupp]{axessibility}  

\usepackage{xcolor}
\usepackage[table]{xcolor}
\usepackage{pifont} 
\definecolor{darkredsoft}{RGB}{210,30,30} 
\newcommand{\redcross}{{\color{darkredsoft}\ding{55}}} 
\newcommand{\greencheck}{{\color{ForestGreen}\ding{51}}} 

\def\paperID{} 
\def\confName{CVPR}
\def\confYear{2026}

\title{\model: Towards Expressive Interactive Gesture Synthesis}

\author{
M. Hamza Mughal$^{1}$\hspace{1.8em} 
Rishabh Dabral$^{1}$\hspace{1.8em} 
Vera Demberg$^{1,2}$\hspace{1.8em}
Christian Theobalt$^{1,2}$\\
$^1$Max Planck Institute for Informatics, SIC 
\hspace{1em}
$^2$Saarland University\\
{\href{https://vcai.mpi-inf.mpg.de/projects/MIBURI/}{\texttt{vcai.mpi-inf.mpg.de/projects/MIBURI}}}
\vspace{-7pt}
\\
}

\begin{document}
\twocolumn[{ 
\renewcommand\twocolumn[1][]{#1} 
\maketitle 
\begin{center} 
    \vspace{-22pt} 
    \includegraphics[width=0.97\linewidth]{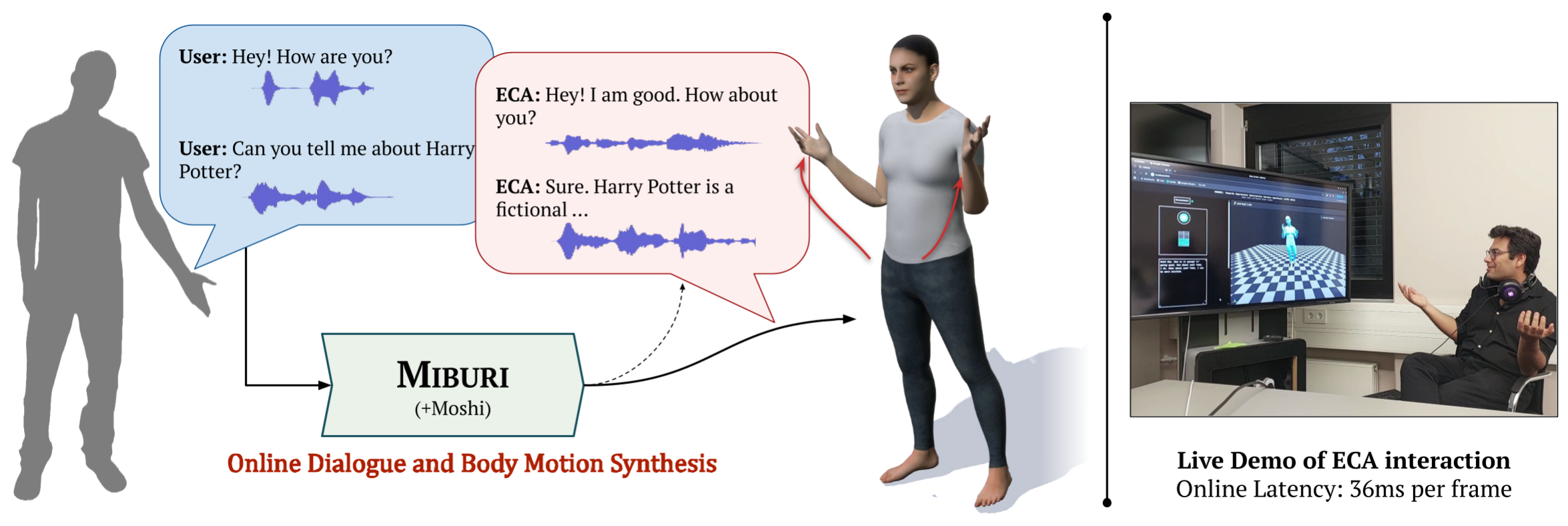}
    \vspace{-12.5pt} 
    \captionof{figure}{\textbf{\model: An online, causal framework for real-time dialogue and gesture generation.}
Given live speech, the system produces full-duplex responses with synchronized full-body gestures. Right: Interactive demo using our approach. 
    }
    \label{fig:TEASER}
\end{center} 
}]
\begin{abstract}
Embodied Conversational Agents (ECAs) aim to emulate human face-to-face interaction through speech, gestures, and facial expressions. 
Current large language model (LLM)-based conversational agents lack embodiment and the expressive gestures essential for natural interaction. 
Existing solutions for ECAs often produce rigid, low-diversity motions, that are unsuitable for human-like interaction.
Alternatively, generative methods for co-speech gesture synthesis yield natural body gestures but depend on future speech context and require long run-times.
To bridge this gap, we present \model, the first online, causal framework for generating expressive full-body gestures and facial expressions synchronized with real-time spoken dialogue. 
We employ body-part aware gesture codecs that encode hierarchical motion details into multi-level discrete tokens.
These tokens are then autoregressively generated by a two-dimensional causal framework conditioned on LLM-based speech-text embeddings, modeling both temporal dynamics and part-level motion hierarchy in real time.
Further, we introduce auxiliary objectives to encourage expressive and diverse gestures while preventing convergence to static poses.
Comparative evaluations demonstrate that our causal and real-time approach produces natural and contextually aligned gestures against recent baselines.
We urge the reader to explore demo videos on~\href{https://vcai.mpi-inf.mpg.de/projects/MIBURI/}{our project page}.
\end{abstract}    
\section{Introduction}
\label{sec:intro}
Human Computer Interaction has evolved from punch card based interfaces to LLM-driven conversational agents. 
Throughout this journey, these interfaces have progressed to emulate a more ``human'' way of interaction. 
Current textual chat assistants are the latest iteration in this evolution, which feature a strong understanding of linguistically encoded world knowledge.
We interact with these digital assistants naturally through our voice or text, without the need to navigate a Graphical User Interface. 
However, human communication is not limited to verbal interaction but also involves non-verbal elements, such as body gestures and facial expressions, which are non-existent in these assistants.
Full-body gestures not only convey meaningful contextual information in a conversation but also structure human interactions, serving as another important means of communication.
\par
Introducing this new communication channel in digital assistants paves the way for \textit{Embodied Conversational Agents}~\cite{cassell1998eca}: interfaces that are more interactive and natural for human communication~\cite{salem2011friendly}, marking a step toward a deeper understanding of the physical world knowledge beyond language.
To achieve this goal of interactive agents, the seminal work of Cassell~\etal~\cite{cassell1998eca} outlines architectural requirements specifying that the system should produce \textit{expressive} body gestures alongside spoken dialogue in \textit{real-time}.
Building on this foundation, both early rule-based~\cite{cassell1998eca, bickmore2005social} and recent data-driven~\cite{Nagy2021gesturebot} approaches have attempted real-time gesture generation synchronized with speech. 
However, they often yield less expressive, low-diversity motion and exhibit artificial turn-taking interaction patterns with distinct speaking and listening phases.
\par
In contrast, recent generative approaches~\cite{ao2023gesturediffuclip, mughal2024convofusion, mughal2025raggesture, zhang2024semanticgesture} produce more natural and expressive co-speech gestures, leveraging neural architectures through diffusion~\cite{ng2024audio2photoreal, diffugesture} or masked modeling in transformers~\cite{liu2024emage}.
However, these models typically operate in an \textit{offline}, non-causal manner, requiring access to both past and future speech context to synthesize motion for a given time step, and thus cannot run in parallel with live speech generation.
It is important to note that \textit{causal} and \textit{real-time} processing are related but distinct requirements: causal models, such as autoregressive transformers, rely only on past inputs, with no regard to any latency requirement, whereas real-time {interactive} systems must additionally meet strict time constraints to maintain conversational fluidity along with expressive gestures.
Consequently, existing generative gesture approaches, while expressive, cannot be used as plug-and-play solution to build the embodied agents {outlined by Cassell \etal~\cite{cassell1998eca}}.
\par
\begin{figure}
    \centering
    \includegraphics[width=\linewidth]{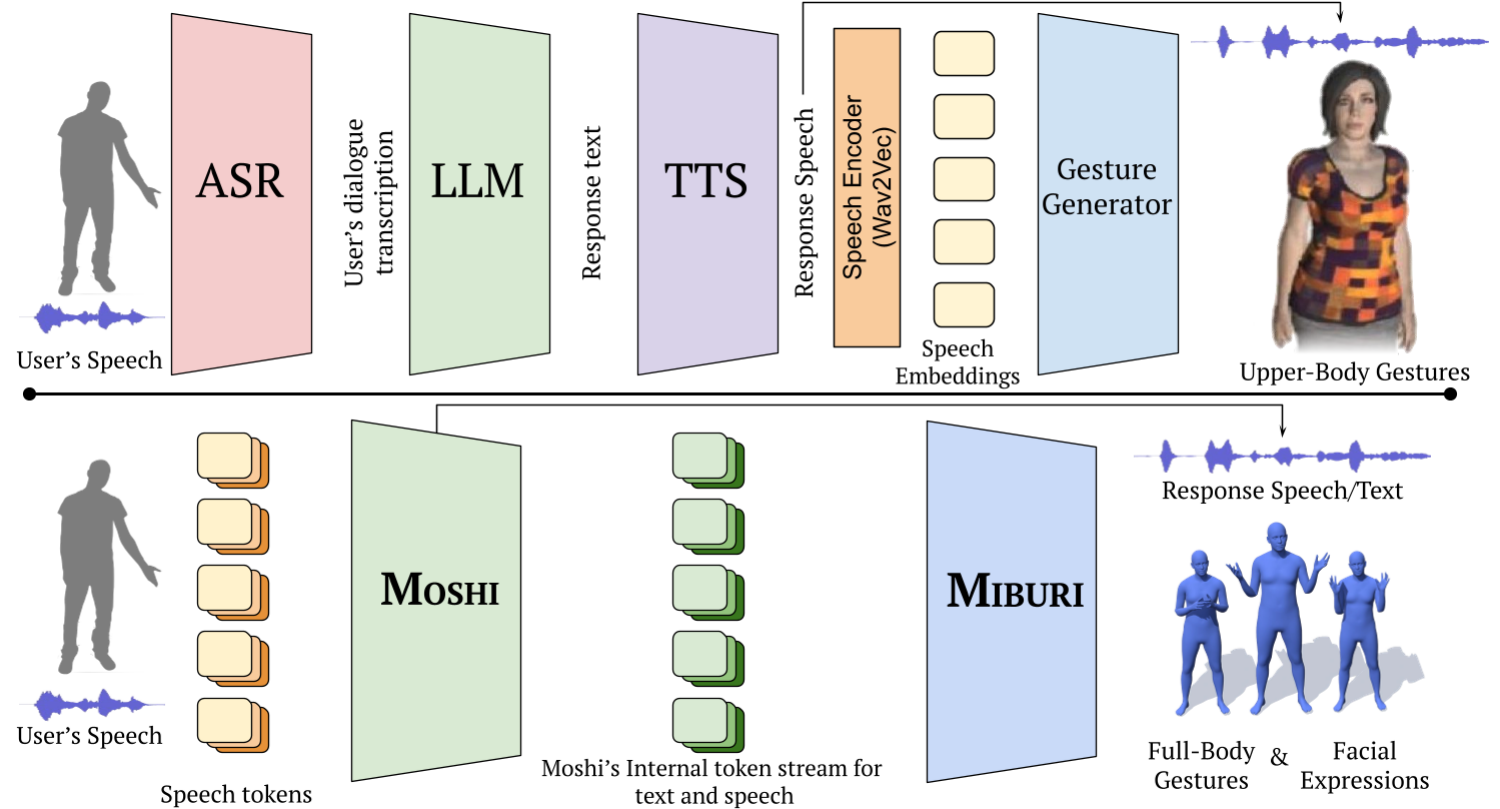}
    \caption{
    \textbf{Overview.} 
    Existing solutions~\cite{Nagy2021gesturebot, chen2025taoavatar} to animate ECAs involve a complex pipeline (above) of multiple components to generate gestures with speech. \model~(below) generates full body co-speech gestures directly by utilizing internal semantic/acoustic tokens of speech-text foundation model~\cite{defossez2024moshi}.}
    \vspace{-5pt}
    \label{fig:overview}
\end{figure}

To address this gap, we introduce \model -- an online, fully causal generative framework that generates expressive co-speech body gestures and facial expressions along with spoken dialogue in real-time.
We build this framework upon Moshi~\cite{defossez2024moshi}, a speech-text foundation model that generates full-duplex spoken dialogue, and leverage its rich contextual speech-text embeddings to generate synchronized body motion.
While several LLM-based gesture synthesis approaches exist~\cite{chen2025taoavatar, mughal2025raggesture}, they typically involve a bulky pipeline in which the LLM outputs are converted to speech, which is then tokenized to condition the gesture synthesis model (\cref{fig:overview} top).
We propose an alternative paradigm.
In order to be causal and real-time, we exploit the speech-text aligned token stream of Moshi, and build our gesture synthesis architecture by directly tapping-on to the internal Moshi tokens.
As illustrated in~\cref{fig:overview} (bottom), this allows us to avoid the latency-inducing steps of the conventional pipelines while benefiting from the rich semantic and acoustic contexts provided by the token embeddings.
\par
Architecturally, our causal generative network leverages these internal tokens to generate gestures through two transformers: one incorporating the \textit{temporal} context and the other generating per-frame, skeleton-aware \textit{kinematic} features.
To facilitate such decomposition, we propose a two-dimensional gesture encoding through Residual VQ-VAE, which is trained to perform causal decoding of the generated gesture tokens.
Noteworthy is that our tokens encode a short temporal window (2 frames) in order to keep the latency low.
We encode gestures by dividing the body into three groups (face, upper and lower body) and tokenize them seperately through individual codecs.
%
\par
In summary, our contributions are threefold:
\begin{itemize}
    \item We contribute a new paradigm for online, real-time and causal gesture generation, which leverages the internal token-stream of a speech-based Large Language Model to perform interactive gesture synthesis.
    \item We carefully design the network architecture and tokenization strategy, which facilitate causal gesture synthesis without compromising the expressiveness of the generated gestures.
    \item We present a comprehensive analysis of the several design choices involved in our method. Through perceptual and numerical experiments, we demonstrate that \model{} advances the state-of-the-art of Embodied Conversational Agents (ECAs).
\end{itemize}
%
\par

\section{Related Work}
\label{sec:related-works}
We first review works on co-speech gesture synthesis, followed by methods for building Embodied Conversational Agents. Although both aim to generate co-speech gestures, their differing requirements make bridging the two fields together non-trivial.
\subsection{Co-Speech Gesture Synthesis}
Co-speech gestures are speech-synchronized body and hand movements that convey semantically aligned meaning~\cite{mcneill1992handmind}.
Existing works on gesture generation range from early rule-based systems~\cite{cassell2001rulebeat, thiebaux2008rulesmartbody, wagner2014rulebasedsurvey} to modern learning-based systems~\cite{ferstl2020adversarial, kucherenko2019analyzing, ghorbani2022zeroeggs, habibie21learning, yoon2019robots}. 
Learning based methods~\cite{liu2024emage, ao2023gesturediffuclip, mughal2024convofusion, ng2024audio2photoreal} are typically data-driven and employ deep networks to convert speech input into synchronized natural-looking motion.
CaMN~\cite{liu2022beat} and EMAGE~\cite{liu2024emage} introduce large-scale speech-aligned motion datasets and transformer-based gesture synthesis methods.
GestureDiffuCLIP~\cite{ao2023gesturediffuclip} utilizes diffusion transformers with causal attention over past and future speech frames.
ConvoFusion~\cite{mughal2024convofusion} uses diffusion to generalize generation across single- and two-person interactions, while Audio2Photoreal~\cite{ng2024audio2photoreal} also generates dyadic interactions with photorealistic avatars.
RAG-Gesture~\cite{mughal2025raggesture} and SemanticGesticulator~\cite{zhang2024semanticgesture} develop retrieval-based paradigms to improve the semantic alignment in generated gestures.
These methods cannot run in real time due to heavy computation, making them unsuitable for online gesture synthesis.
\par
To address long runtimes, MambaTalk~\cite{xu2024mambatalk} uses selective state-space models with non-causal cross-attention for low-latency generation.
GestureLSM~\cite{liu2025gesturelsm} tackles this with a real-time flow-matching framework and shortcut sampling.
Both methods also require seed gesture sequences during inference.
However, these methods remain offline and non-causal, relying on past and future speech, and therefore cannot support online ECAs. 
This highlights the need for a real-time, causal framework that generates expressive gestures directly from speech without future context, seed gestures, or long runtimes.
%
\subsection{Embodied Conversational Agents (ECAs)}
In language generation, LLMs~\cite{touvron2023llama, openai2024gpt} have shown strong capability to generate and understand text.
Similarly, recent spoken dialogue systems~\cite{defossez2024moshi, openai2024gpt4ocard} aim to perform conversations in real-time while maintaining knowledge and reasoning abilities exhibited by LLMs. 
However, these natural language interfaces lack the full-body dynamics for an embodied avatar.
In the avatar space, recent approaches have tried to enhance LLM-driven conversations with virtual characters through articulated body movements.
Digital Life Project~\cite{cai2024dlp} uses an LLM backbone to synthesize instruction-driven motion for virtual characters.
TaoAvatar~\cite{chen2025taoavatar} focuses on producing a full-body photorealistic avatar in real-time, given gesture input from motion library.
\par
Full-fledged solutions for ECAs mainly include rule-based systems, while there are only a few recent data-driven systems.
Rule-based systems~\cite{cassell2000embodied, cassell1998eca, bickmore2005social} usually utilize pre-recorded animations to synthesize body motion in real time.
Hybrid systems~\cite{llanes2024developing, wampfler2025platform} use neural methods for lip animation synthesis and a rule-based approach for body gestures.
Recently, Abel~\etal~\cite{abel2024towards} propose a GRU-based pipeline to generate real-time co-speech gestures.
Nagy~\etal present Gesturebot~\cite{Nagy2021gesturebot} that utilizes data-driven methods like~\cite{kucherenko2020gesticulatormlp} to create an embodied avatar for body gestures.
However, Gesturebot is limited to manual turn-based interactions and animates gestures only during speech, using a non-causal model. In contrast, our framework operates causally at both speech and gesture token levels, enabling real-time, continuous interaction.
\begin{table}[]
\centering
\caption{Approaches for offline Gesture Synthesis and ECAs}
\resizebox{\linewidth}{!}{%
\begin{tabular}{lcccc}
\textbf{Method} & \textbf{Approach} & \textbf{Expressive} & \textbf{Causal} & \textbf{Real-time} \\ \hline
Cassell et. al.~\cite{cassell1998eca, cassell2000embodied}  & Rule-based               & \redcross   & \greencheck & \greencheck \\
DigitalEinstein~\cite{wampfler2025platform} & Rule-based               & \redcross   & \greencheck & \greencheck \\
Gesturebot~\cite{Nagy2021gesturebot}              & Autoregressive           & \redcross   & \redcross & \greencheck \\ \hline
EMAGE~\cite{liu2024emage}                   & Masked Gesture Modelling & \greencheck & \redcross   & \redcross   \\
ConvoFusion~\cite{mughal2024convofusion}            & Diffusion                & \greencheck & \redcross   & \redcross   \\
Audio2Photoreal~\cite{ng2024audio2photoreal}         & VQ+Diffusion             & \greencheck & \redcross   & \redcross   \\
RAG-Gesture~\cite{mughal2025raggesture}             & Retrieval+Diffusion      & \greencheck & \redcross   & \redcross   \\
GestureLSM~\cite{liu2025gesturelsm}              & Flow-Matching            & \greencheck & \redcross   & \greencheck \\
MambaTalk~\cite{xu2024mambatalk}               & SSM                      & \greencheck & \redcross   & \greencheck \\ \hline
\model~(Ours)                    & RVQ+Autoregressive       & \greencheck & \greencheck & \greencheck \\ \hline
\end{tabular}
}
\label{tab:method-comparison}
\end{table}

\begin{figure*}
    \centering
    \includegraphics[width=0.95\linewidth]{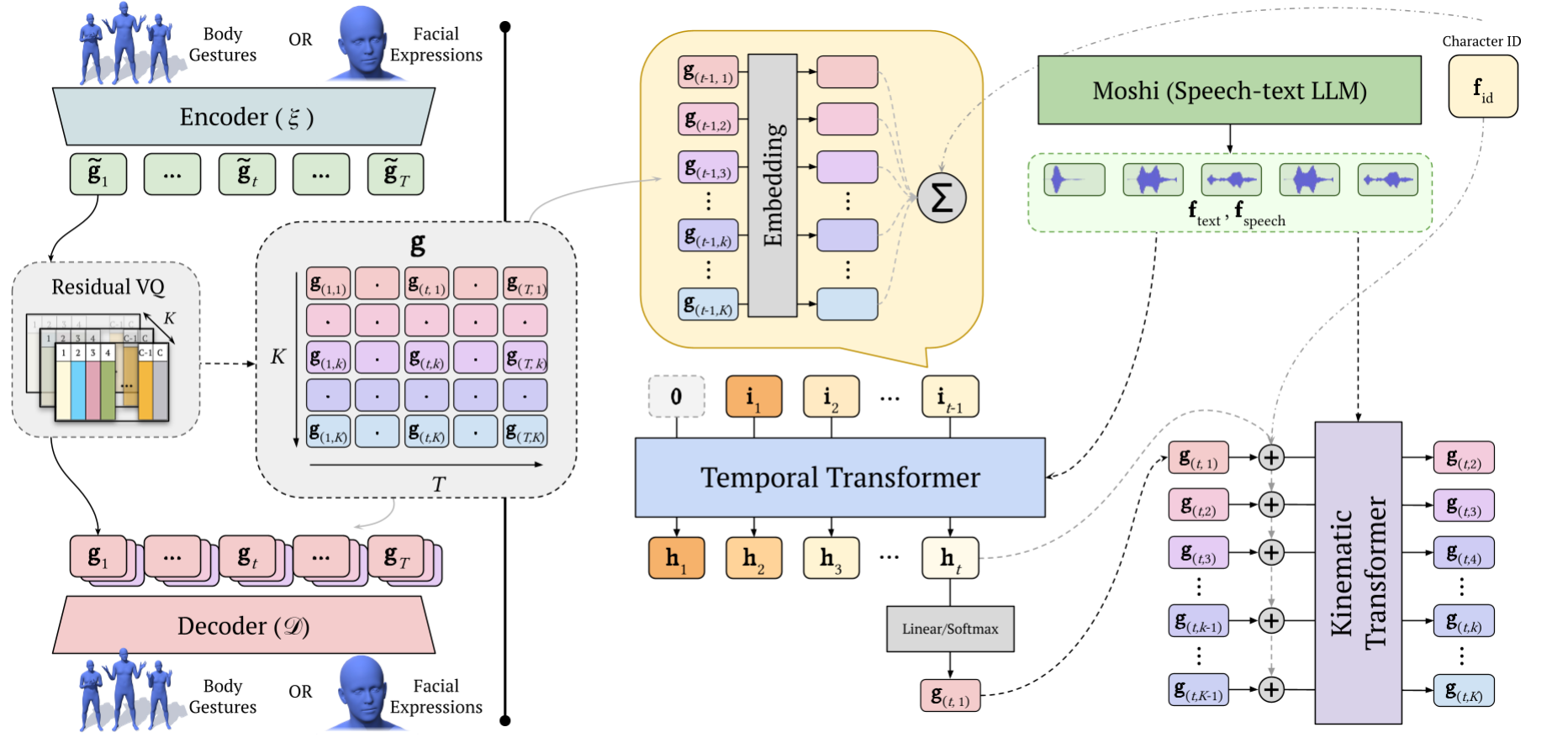}
    \caption{\textbf{\model~Architecture.} Given Moshi's speech/text tokens(\cref{subsec:moshi}), our approach generates a sequence of gesture tokens, which are obtained through Body-part aware Gesture Codecs(\cref{subsec:rvq}). This online framework takes in Moshi's text/speech token as input and predict gesture tokens through autoregressive \textit{temporal} and \textit{kinematic} transformers(\cref{subsec:tdm}).
    \vspace{-10pt}
    }
    \label{fig:approach}
\end{figure*}
\section{Approach}
\label{sec:method}
The goal of our approach is to generate full-body gestures and facial expressions synchronized with speech for ECAs.
To enable such interactive agents, an ideal framework must produce spoken dialogue and then leverage the underlying verbal and prosodic context to synthesize \textit{expressive} and diverse body gestures.
According to the Cassell et al.~\cite{cassell1998eca}, there are two key requirements for interactive gesture synthesis:
(1) it must be \textit{causal}, $i.e.$ one cannot assume the availability of future utterance,
(2) and it must be \textit{realtime} with low latency.
Here, it is important to emphasize that simply having a low amortized rate of generation, as is typical for diffusion-based methods, is not enough.
\par
Our approach builds upon a speech-text foundation model~\cite{defossez2024moshi} to generate full-duplex spoken dialogue and extract its internal speech-text token stream to provide rich contextual input for gesture synthesis (\cref{subsec:moshi}).
Our gesture generator then autoregressively produces body-region aware motion tokens (\cref{subsec:rvq}) using two-dimensional temporal and kinematic transformers(~\cref{subsec:tdm}). 
\par
However, this base generation framework is insufficient to achieve diverse and expressive body gestures, which are crucial for natural communication.
Therefore, we propose additional objectives for our autoregressive framework to achieve human-like gesture quality (\cref{subsec:losses}).
Finally, we ensure causal and real-time inference by carefully designing attention contexts and efficient cache mechanisms (\cref{subsec:implementation}).
\cref{fig:approach} illustrates our proposed architecture.
\subsection{Preliminaries: Moshi}
\label{subsec:moshi}
To produce real-time conversational speech and language, we utilize an open-source spoken dialogue system, i.e. Moshi~\cite{defossez2024moshi}.
Built on a textual LLM backbone, this framework autoregressively generates text and speech tokens. 
Crucially, it enables full-duplex conversations by jointly modeling its own speech and the user's speech in parallel token streams.
At output, it generates speech and text tokens i.e. $\mathbf{f}^{\text{speech}} \in \mathbb{R}^{T \times K^{\text{speech}} \times d}$ and $\mathbf{f}^{\text{text}} \in \mathbb{R}^{T \times K^{\text{text}} \times d}$, where $T$ represents the number of tokens along the time axis and $d$ denotes the embedding dimension. 
Moshi also utilizes Residual Vector Quantization~\cite{zeghidour2021soundstream} to encode speech into multiple levels of semantic and acoustic tokens, and $K^{\text{speech}}$ and $K^{\text{text}}$ represent those quantization levels for speech and text respectively.
\model~aims to leverage these semantic and prosodic details from Moshi for generating its own skeleton-aware token stream of gestures.
\subsection{Body-part wise Gesture Codecs}
\label{subsec:rvq}
As the first step in our framework, we build a robust motion prior that encodes gesture frames into discrete tokens, which can then be used for downstream gesture generation using an autoregressive transformer.
Since co-speech articulation in each body region happens at different scales~\cite{mughal2024convofusion} and different body regions relate to speech seperately~\cite{liu2024emage, mughal2025raggesture}, we decouple each pose in the gesture sequence $\mathbf{x}$ into three body regions: 
upper body with hands $\mathbf{x}^u \in \mathbb{R}^{N \times 6J^{u}}$, 
lower body with global translation and foot contacts $\mathbf{x}^l \in \mathbb{R}^{N \times (6J^l+3+4)}$, and 
facial expressions using FLAME parameters $\mathbf{x}^f \in \mathbb{R}^{N \times (100+6J^f)}$~\cite{FLAME:SiggraphAsia2017}.
Here $N$ is the number of frames of human motion, while $J^{u}$, $J^l$ and $J^f$ represent upper body, lower body and jaw joints respectively~\cite{SMPL-X:2019, zhou2019continuity}. 
Each region-specific gesture sequence is encoded through a separate Gesture Codec, which utilizes Residual VQ-VAE for motion tokenization.
\paragraph{Residual VQ-VAE for Gestures.}
Co-speech body articulation contains multiple aspects of detail, ranging from large arm jerks to subtle finger-level gestures.
Na\"ively tokenizing gestures through VQ-VAE quantization schemes~\cite{liu2024emage, suresh2025modeling, suresh2025enhancing} can result in coarse and choppy motion due to the loss of finer kinematic details (see \cref{subsec:ablation}).
To encapsulate these subtle motion details, we train gesture codecs for each body region using Residual VQ-VAE~\cite{zeghidour2021soundstream}. 
Each region-wise codec consists of an encoder-decoder architecture (\cref{fig:approach}) with encoder $\mathcal{E}^{b}$ containing downsampling 1-d convolution layers and a transformer encoder with causal self-attention.
It encodes motion for a given body region as $\mathbf{\tilde{g}}^b = \mathcal{E}^{b}(\mathbf{x}^b)$, whose output is quantized into tokens $\mathbf{{g}}^b \in \mathbb{R}^{T \times K^b}$ with $K^b$ levels of motion details via Residual Vector Quantization (RVQ).
Here $T$ is the temporal length of the token sequence, downsampled from $N$.
Each residual level learns a codebook $\mathbf{C}_k \in \mathbb{R}^{V \times d}$ that is used for vector quantization of the corresponding residual.
Consequently, motion can be reconstructed through the decoder: $\mathbf{\hat{x}}^b = \mathcal{D}^{b}(\mathbf{{g}}^b)$, which consists of a similar causal transformer encoder and upsampling transpose 1D-convolution layers.
\par
These gesture tokenizing codecs are trained with a set of reconstruction/geometric losses and a latent embedding loss at each quantization level (detailed in the supplemental).
Finally, the resulting gesture sequence $\mathbf{{g}} \in \mathbb{R}^{T \times K}$ can be defined as a concatenation
:
\[
\mathbf{{g}} = \operatorname{Concat}(
\mathbf{{g}}^u, \mathbf{{g}}^l, \mathbf{{g}}^f)
\]
along the $K$ level axis, with $K = K^u + K^l + K^f$.
This tokenized sequence $\mathbf{{g}} = \{\mathbf{{g}}_{(t, k)} \mid t = 1 \dots T,, k = 1 \dots K \}$ represents motion along \textit{temporal} and \textit{kinematic} dimensions, where  
former encompasses kinematic details across time and the latter contains part-level details across body regions at each $t$.
\subsection{Autoregressive \& Causal Transformers}
\label{subsec:tdm}
Recall that our objective is to design a \textit{causal} gesture synthesis framework, which needs to generate gesture tokens $\mathbf{{g}}$, given speech $\mathbf{f}^{\text{speech}}$ and text $\mathbf{f}^{\text{text}}$ tokens from Moshi and a character identity embedding $\mathbf{f}^{\text{id}}$ as input.
Autoregressive transformers are commonly used in causal next-token prediction tasks, where attention layers attend to the previous $T$ tokens. 
However, in our case, each token frame in $T$ also contains $K$ token levels representing hierarchical motion details. 
A na\"ive implementation would require us to model $T.K$ tokens autoregressively, where attention layers would need the context of at least $>K$ tokens to learn temporal dynamics across motion frames.
This automatically increases the size of context window in attention layers, while being harder to train and computationally expensive at inference (see \cref{subsec:ablation}). 
Therefore, we disentangle the prediction of both temporal and kinematic dimensions of gesture codecs $\mathbf{{g}}$ with two transformers inspired by RQ-Transformer~\cite{lee2022autoregressive, zhu2024generative, defossez2024moshi}.
\paragraph{Temporal Transformer.}
First, we build our base \textit{temporal} transformer $\mathcal{T}_{\text{temporal}}$ to focus on the temporal dynamics across time.
This causal transformer is trained to autoregressively predict the first level token $\mathbf{{g}}_{(t,1)}$ (among the $K$ kinematic levels), conditioned on the tokens from previous timesteps.
\begin{align}
\mathbf{h}_{t} &= \mathcal{T}_{\text{temporal}} \big(\mathbf{{g}}_{(<t)}, \mathbf{f}^{\text{speech}}_{(\leq t)}, \mathbf{f}^{\text{text}}_{(\leq t)}, \mathbf{f}^{\text{id}} \big) \\
\mathbf{{g}}_{(t,1)} &= \text{Softmax} \big( \text{Linear}(\mathbf{h}_{t}) \big)
\end{align}
Internally, the embeddings for $\mathbf{{g}}_{(<t)}$ along the kinematic dimension $K$ are summed up to form a single input $\mathbf{i}_{t-1}$ for each $t$ (see \cref{fig:approach}).
The output of transformer $\mathbf{h}_{t}$ is converted to logits $\mathbf{{o}}_{(t,1)} \in \mathbb{R}^{V}$ through a simple classification layer and then we obtain $\mathbf{{g}}_{(t,1)}$ through Softmax.
This module is implemented as a transformer decoder with a causal self-attention over past gesture tokens and dual causal cross-attention layers attending to preceding and current speech and text tokens.
Note that we also learn per-identity feature embeddings that are added at each timestep.
\paragraph{Kinematic Transformer.}
Next, we model the \textit{kinematic} dimension of gesture tokens through a transformer $\mathcal{T}_{\text{kinematic}}$, which autoregressively predicts the next body-part level at each timestep $t$.
In addition to the previously generated levels, the kinematic transformer conditions on the temporal context $\mathbf{h}_{t}$, as well as speech, text, and identity embeddings:
\begin{equation}
    \mathbf{{g}}_{(t,k)} = \mathcal{T}_{\text{kinematic}} \big( \mathbf{h}_{t}, \mathbf{{g}}_{(t,<k)}, \mathbf{f}^{\text{speech}}_{t}, \mathbf{f}^{\text{text}}_{t}, \mathbf{f}^{\text{id}} \big)
\end{equation}
Here, the timestep $t$ remains fixed for each level prediction.
Therefore, the speech and text inputs correspond only to embeddings at time $t$.
This transformer is also implemented as a decoder with a causal self-attention layer and cross-attention layers for speech and text.
The identity embedding and temporal context $\mathbf{h}_{t}$ are added to the input of each level-step.
Finally, the output of each step $k$ is projected through classification layers to predict $\mathbf{{g}}_{(t,k)}$.
Note, that at the first level-step, the $\mathcal{T}_{\text{kinematic}}$ receives $\mathbf{{g}}_{(t,1)}$ from the temporal transformer as input and predicts the next level $\mathbf{{g}}_{(t,2)}$ and further.
\par
We jointly train both transformers using cross-entropy loss $\mathcal{L}_{\text{CE}}$ over the ground-truth tokens and employ teacher-forcing during the training process.
\subsection{Improving Expressiveness}
\label{subsec:losses}
Autoregressive architectures for motion synthesis excel at generating coherent motion sequences, especially in causal scenarios. 
However, they tend to converge to mean-poses and accumulate drifts along the temporal dimension~\cite{dabral2023mofusion, kucherenko2020gesticulatormlp}.
To obtain expressive gestures, we introduce auxiliary objectives that explicitly encourage motion diversity and prevent collapse into static or repetitive gestures.
\par
During training, we apply a contrastive InfoNCE loss~\cite{oord2018representation} over the predicted tokens to improve gesture expressiveness.
However, sampling from a discrete distribution is non-differentiable and will not allow this loss to contribute during training.
Hence, we resort to the Gumbel-Softmax reparameterization trick~\cite{jang2016categorical} to approximate the discrete sampling process.
This allows us to obtain probabilities from the logit outputs $\mathbf{\tilde{o}} \in \mathbb{R}^{{T \times K \times V}}$ of temporal and kinematic transformers, which are then converted to the latent output of RVQ step (from \cref{subsec:rvq}):
\begin{equation}
\mathbf{z} = \sum_{k=1}^{K} \text{GumbelSoftmax}(\mathbf{\tilde{o}}_{k}) \mathbf{C}_{k} \in \mathbb{R}^{T \times d}
\label{eq:gumbel_rvq}
\end{equation}
We use GumbelSoftmax with a temperature of $0.4$ and sample one-hot vectors at its output using differentiable straight-through estimator.
The latent output $\mathbf{z}$ is calculated separately for each body region by using its corresponding RVQ codebooks.
Given ground-truth latents $\mathbf{z}^{\text{GT}}$ and generated latents $\mathbf{z}^{\text{pred}}$, we compute a similarity matrix across all real-fake pairs and apply an Info-NCE loss:
\begin{equation}
\mathcal{L}_{\text{con}} =
- \mathbb{E}_{i} \Bigg[
\log \frac{
    \exp(\text{sim}(\mathbf{z}_i^{\text{GT}}, \mathbf{z}_i^{\text{pred}}) / \tau)
}{
    \sum_{j=1}^{B} \exp(\text{sim}(\mathbf{z}_i^{\text{GT}}, \mathbf{z}_j^{\text{pred}}) / \tau)
}
\Bigg].
\end{equation}
where $\text{sim}(\cdot)$ denotes cosine similarity and $\tau$ is the temperature parameter.
This loss enforces high similarity between matching GT–predicted latents while pushing apart mismatched pairs across the batch $B$, resulting in more expressive and speech-aligned motion generation. 
\par
In practice, we apply this loss over temporal segments of $\mathbf{z}$ instead of the complete temporal length $T$, in order to encourage similarity in motion trajectories across gesture phases.
\paragraph{Voice Activation Loss.} 
Our framework generalizes to both listening and speaking states of body gestures. 
Since, humans gesticulate differently while listening or speaking, we explicitly enforce our network to learn the distinction between the two states. 
This is achieved by projecting the transformer output $\mathbf{h}_t$ onto a binary-classification head that classifies $\mathbf{h}_t$ into listening (0) or speaking (1) states.
Trained with a Binary Cross-Entropy loss $\mathcal{L}_{\text{va}}$, this auxiliary task head prevents \textit{phantom} gestures during the listening state and forces speech-aligned expressive gestures during the speaking stage.
\par
Finally, the complete network is optimized through joint loss $\mathcal{L} = \mathcal{L}_{\text{CE}} + \alpha \mathcal{L}_{\text{con}} + \beta \mathcal{L}_{\text{va}}$, with $\alpha, \beta$ being loss weights.

\subsection{Implementation}
\label{subsec:implementation}
To enable gesture generation in real-time that is time-aligned with Moshi, we implement efficient techniques to achieve faster synthesis times.
Moshi's latency is 200ms at a rate of 12.5 tokens per second, where each token represents 0.08 seconds of audio, and hence, \model~also generates 0.08 seconds of gestures and facial expressions at each timestep.
Our training data contains 25 FPS motion, which means our framework generates 2 frames at each step.
Gesture codecs contain $K_u=K_l=8$ and $K_f=4$ residual levels and we use $T=125$ during training which amounts to a 10-second motion sequence.
Temporal transformer consists of 4 layers with 2 attention heads and the kinematic transformer consists of 2 layers and 1 attention head.
Training optimization is done using AdamW~\cite{loshchilov2017decoupled} with starting learning rate of 1e-4, which is annealed across epochs.
\par
For efficient attention inference, we store key and values for previous timesteps in a KV-Cache to retain the context required during attention.
We limit the attention context of self-attention layers to 25 tokens and keep a longer context of 50 tokens for cross-attention layers with speech and text.
The temporal transformer starts inference with a zero initial token to predict $\mathbf{{g}}_{(1,1)}$ (refer to \cref{fig:approach}).
In practice, due to the small relationship between the lower body and speech/text, we mask out cross-attention for lower body tokens to save runtime.
During training, we set $\alpha$ and $\beta$ to 0.1 and 0.01, respectively.
At inference time, we generate tokens using top-p (nucleus) sampling~\cite{holtzman2019curious}, instead of greedy sampling, to maintain diversity. We set top-p for the temporal transformer to 0.8 and for the kinematic transformer to 0.95, with the softmax temperature of 0.9 for both.
Moreover, we apply classifier-free guidance (CFG)~\cite{ho2022classifier} during sampling to improve gesture alignment with Moshi's rich semantic and acoustic information.

\section{Experiments}
\label{sec:experiments}
We evaluate our approach against state-of-the-art baselines for co-speech gesture synthesis.
We perform perceptual (\cref{subsec:userstudy}) and quantitative (\cref{subsec:metrics}) evaluations to measure gesture quality, motion naturalness and speech appropriateness.
Moreover, we also analyze generation times for each baseline to measure real-time capability.
Lastly, we validate our design choices through ablative analysis (\cref{subsec:ablation}).
\par
Baseline methods include two types of approaches: (1) Non-causal and non real-time approaches like RAG-Gesture~\cite{mughal2025raggesture}, EMAGE~\cite{liu2024emage} and CaMN~\cite{liu2022beat}, which aim to synthesize meaningful expressive motion, and (2) Real-time approaches like GestureLSM~\cite{liu2025gesturelsm} and MambaTalk~\cite{xu2024mambatalk}, which have fast sampling times during generation.
Since there are no causal neural baselines, we also implement causal versions of real-time methods~\cite{liu2025gesturelsm, xu2024mambatalk} to compare our approach with na\"ive implementations of causal gesture synthesis (details in supplemental).
It is important to note that all baselines (except \cite{mughal2025raggesture}) require a seed sequence and leverage its context to generate motion, whereas our framework does not.
\paragraph{Dataset.}
We train our approach on the BEAT2 dataset~\cite{liu2024emage} and evaluate its performance on standard train/val/test split from the dataset.
The dataset originally contains 25 speakers, but we remove 2 speakers (\textit{carla} \& \textit{itoi}) to ensure good quality motion-tracking for our training/evaluation data.
Following~\cite{mughal2025raggesture} and unlike other baseline BEAT2 methods, we evaluate on both single-speaker (\textit{scott}) and multi-speaker test sets to assess performance on large-scale multi-speaker setting. Our test set contains 15 and 249 full-length utterances for 1-speaker and 23-speaker setting respectively.
We retrain baselines for the multi-speaker setting if their multi-speaker variant is unreleased.
Lastly, we also provide an evaluation on the recently released Embody3D dataset~\cite{embody3d} in the supplemental material.
\subsection{Perceptual Evaluation}
\label{subsec:userstudy}
\begin{figure}
    \centering
    \includegraphics[width=\linewidth]{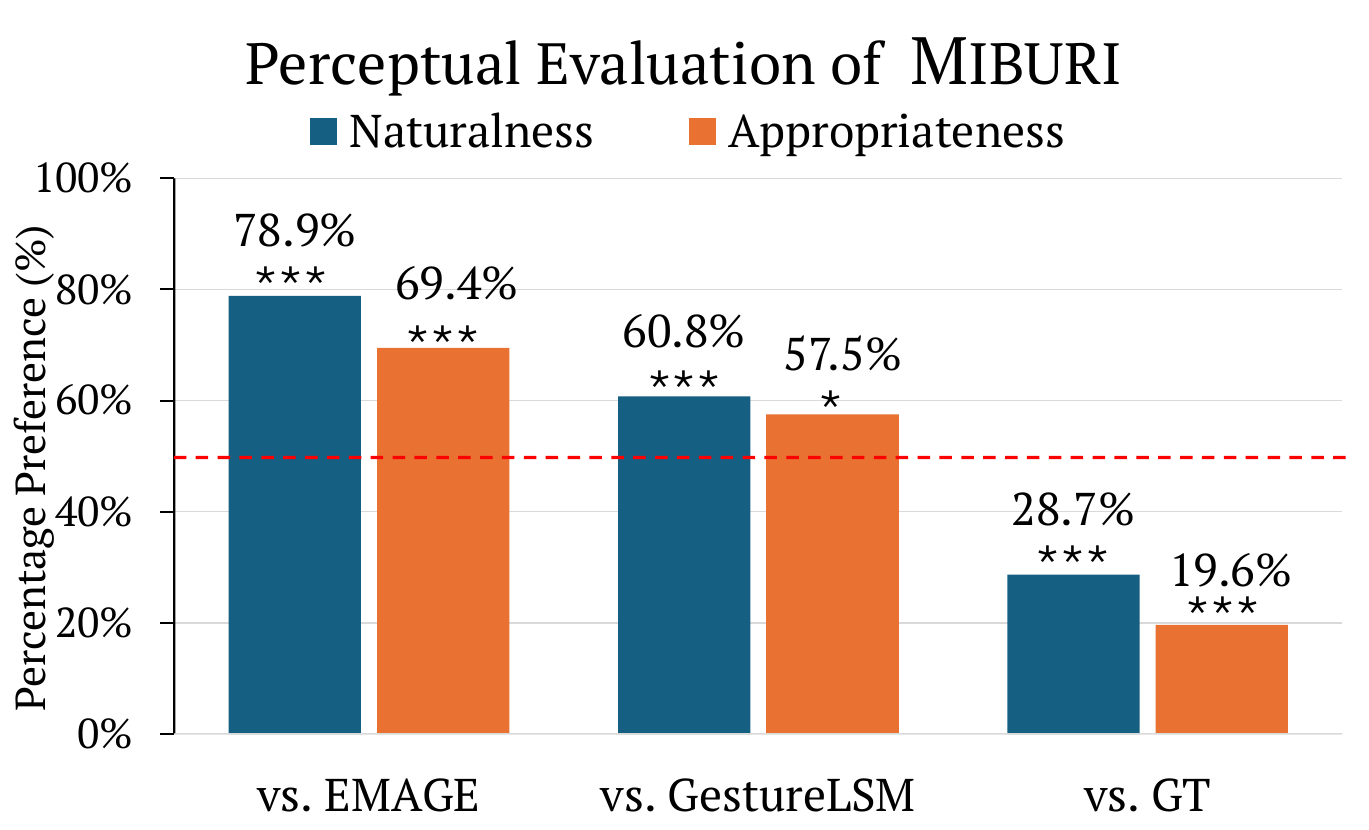}
    \caption{
    \textbf{User Study for Perceptual Evaluation.} Here, the red line indicates chance level (50\%), * stands for $p < 0.05$ and *** for $p < 0.001$.}
    \vspace{-10pt}
    \label{fig:user-study}
\end{figure}

Quantitative metrics focus on singular aspects of the gesture generation problem, and have yet to represent correlation with human perception of gestures~\cite{nagy2025gesture}. 
%
%
Therefore, we perform a perceptual evaluation on BEAT2 test set to holistically evaluate aspects of gesture synthesis like \textit{naturalness} of motion and \textit{appropriateness} to given speech (\cref{fig:user-study}).
Participants perform pair-wise comparison between \model's gesture outputs and generations from baseline methods.
Results demonstrate \model's ability to generate expressive and natural motion over standard non-causal baselines like EMAGE~\cite{liu2024emage} and GestureLSM~\cite{liu2025gesturelsm}.
However, we observe that our framework has yet to achieve similar quality and speech appropriateness against ground truth data.
Further details are given in the supplementary material.
\subsection{Quantitative Evaluation}
\label{subsec:metrics}
Evaluation metrics for gesture synthesis include Beat-Alignment~\cite{aichoreo}, Frechet Gesture Distance~\cite{yoon2019robots}, L1 Divergence and Diversity.
Each metric aims to measure a specific aspect of gesture quality, with FGD measuring distribution alignment to ground-truth data and BeatAlign gauging prosodic alignment of motion with speech.
To be consistent with BEAT2 baseline methods, we first evaluate our approach on single speaker setting on BEAT2 (\cref{tab:1-spk-eval}) and then perform multi-speaker evaluation across 23 speakers (\cref{tab:all-spk-eval}). 
For single-speaker training data, we observe comparable performance against non-causal baselines in terms of BeatAlign. 
Methods which generate gestures from ground-truth seed sequences understandably perform better in single-speaker setting, achieving lower FID.
We set \model's CFG scale to 1.5 in single speaker setting.
\par
More importantly, our framework achieves state-of-the-art metric performance in FGD and BeatAlign, when trained with a larger number of speakers. 
Firstly, this entails that our causal approach benefits from larger and more diverse motion data and scales well across multiple identities, without the need for seed sequences and future context.
Secondly, when comparing causal versions of existing methods, we find that na\"ively converting baselines to be trained in a causal fashion, leads to worse performance even if the method is real-time.
This also shows current architectures' dependance on future speech context in order to achieve good quality.
Lastly, we also trained larger versions of \model~in this setting to gauge the effect of model sizes without being limited from real-time constraints.
However, we find that leaner versions of \model~are equivalent or better.
We set the CFG scale to 2.3 in multi-speaker setting.

\begin{table}[]
\caption{\textbf{Multi-speaker evaluation.} Facial-MSE scaled by $10^{-8}$. * refers to retrained methods.}
\resizebox{\linewidth}{!}
{%
\begin{tabular}{lcccc}
                        & \multicolumn{4}{c}{Multiple speakers (23)}     \\ \cline{2-5} 
                        & FGD$\downarrow$    & BeatAlign$\rightarrow$    & L1-Div$\rightarrow$ & Facial-MSE$\downarrow$ \\ \cline{2-5} 
GT                      &       & $0.446$ & $8.45$   &            \\ \hline
CaMN                    & $0.736$ & $0.176$ & $6.73$   &  --          \\
EMAGE*                   & $0.850$ & $0.236$ & $6.58$   & $4.6$     \\
RAG-Gesture             & \silver{$0.515$} & $0.648$ & $10.09$  &  --          \\
GestureLSM              & $0.537$ & $0.481$ & \gold{$8.41$}   &   --         \\
GestureLSM (Causal*)    & $2.792$ & $0.684$ & \silver{$9.11$}   &    --        \\
MambaTalk*               & $1.375$ & $0.080$  & $3.73$   &     \gold{$4.12$}       \\
MambaTalk (Causal*)     & $1.222$ & $0.102$ & $4.61$  & \silver{$4.17$}           \\ \hline
\model-L          & $0.555$ & \silver{$0.431$} & $9.45$   &   --         \\
\model-L (+Face)   & $0.582$ & \gold{$0.434$} & $9.31$   & $7.63$    \\ 
\model             & $0.585$ & $0.415$ & $9.75$   &       --     \\
\model (+Face)    & \gold{$0.480$} & $0.461$ & $10.44$  & $7.77$    \\ \hline
\end{tabular}
}
\label{tab:all-spk-eval}
\end{table}
\begin{table}[]
\centering
\caption{\textbf{Single-speaker evaluation.} Facial-MSE scaled by $10^{-8}$.}
\resizebox{\linewidth}{!}{%
\begin{tabular}{lcccc}
                     & \multicolumn{4}{c}{Single-speaker (\textit{Scott})} \\ \cline{2-5} 
                     & FGD$\downarrow$    & BeatAlign$\rightarrow$    & L1-Div$\rightarrow$ & Facial-MSE$\downarrow$ \\ \cline{2-5} 
GT                   &        & $0.749$ & $13.22$  &   --         \\ \hline
CaMN                 & $0.969$  & $0.698$ & $10.61$  &   --         \\
EMAGE                & $0.552$  & $0.795$ & \gold{$13.06$}  & \silver{$7.68$}    \\
RAG-Gesture          & $0.879$  & \gold{$0.730$} & $12.62$  &   --         \\
GestureLSM           & \gold{$0.410$}  & $0.719$ & \silver{$13.42$}  &    --        \\
GestureLSM (+Face)   & \silver{$0.424$}  & \silver{$0.729$} & $13.76$  & $10.20$    \\
MambaTalk            & $0.530$  & $0.779$ & $12.99$  & \gold{$6.25$}    \\ \hline
\model         & $0.806$  & $0.790$ & $17.5$   & --          \\
\model (+Face)  & $0.753$  & $0.790$ & $15.85$  & $8.85$    \\ \hline
\end{tabular}
}
\label{tab:1-spk-eval}
\end{table}
\subsection{Latency Analysis}
\label{subsec:latency}
Recall that having low latency is critical for enabling seamless interactions with the end-user.
Consequently, keeping the latency low has been one of the key design considerations in our method.
Our online demo system achieves a latency of \textbf{36ms per frame} on RTX3090.
This includes model's runtime and rendering on a web dashboard (see Suppl. Mat.).
Moreover, we present a comparative analysis of \model's latency with respect to existing state-of-the-art methods in~\cref{tab:latency}.
Having a low token context (2 frames) helps our autoregressive design and we achieve the lowest latency.
In contrast, non-autoregressive diffusion-based methods need to wait for all the context-frames to be generated in order to render the output, thereby leading to high latency.
Interestingly, while MambaTalk~\cite{xu2024mambatalk} is based on the inherently causal Mamba~\cite{mamba} architecture, their decision to inject speech conditioning through a cross-attention layer becomes counter-productive for generating low-latency outputs.
Our proposed \model{} architecture strikes a balance between gesture quality and generation latency.

\begin{table}[t]
\centering
\caption{\textbf{Latency and Causality Comparison.} Wall-clock time is measured from the beginning of the forward pass to the conversion of outputs into SMPL-X parameters. Render times are excluded here.
\#Frames / Step indicates the number of frames generated per forward pass.}
\resizebox{\linewidth}{!}{%
\begin{tabular}{lccc}
\toprule
 & Causal 
 & $\text{Latency}_{\text{A100}}\!\downarrow$
 & \#Frames / Step \\
\midrule
GestureLSM (8 steps)
& \redcross
& $0.1447 \pm 0.0034$
& $124$ \\

EMAGE
& \redcross
& $0.0374 \pm 0.0004$
& $60$ \\

MambaTalk
& \redcross
& $0.0529 \pm 0.0039$
& $60$ \\

\textbf{\model~(ours)}
& \greencheck
& $0.0349 \pm 0.0017$
& $2$ \\
\bottomrule
\end{tabular}
}

\label{tab:latency}
\end{table}

\subsection{Ablation Studies}
\label{subsec:ablation}
We perform ablative analysis over different aspects of our framework, ranging from choice of speech encodings, architecture/loss design and motion tokenization strategy. 
\paragraph{Comparison of Speech/Text Encodings.}
Since existing systems utilize a multi-step pipeline to generate body gestures in ECAs (\cref{fig:overview}), we analyze the most important part of that pipeline for gesture synthesis i.e. speech input encoding, and compare it with our approach of leveraging Moshi's internal token stream.
We compare the performance of our gesture synthesis model \model, by training it with internal embeddings of Moshi tokens and also, by using standard wav2vec~\cite{baevski2020wav2vec} based encoding, which is common in gesture synthesis frameworks~\cite{mughal2025raggesture}.
\begin{table}[t]
\centering
\caption{\textbf{Wav2vec ablation against moshi features.}}
\resizebox{\linewidth}{!}{%
\begin{tabular}{lccc}
\toprule
 & $\text{FGD}\!\downarrow$ & $\text{BeatAlign}\!\rightarrow$ & $\text{L1-Div}\!\rightarrow$ \\
\midrule
GT                        
& --      & 0.446 & 8.45 \\ 
\midrule
\model-L (+Face) w/ wav2vec 
& 0.595   & 0.404 & 7.92 \\

\model (+Face) w/ wav2vec  
& 0.665   & 0.363 & \silver{7.07} \\
\midrule
\model-L (+Face)            
& \silver{0.582} & \gold{0.434} & \gold{9.31} \\

\model (+Face)              
& \gold{0.480} & \silver{0.461} & 10.44 \\
\bottomrule
\end{tabular}
}

\label{tab:wav2vec-eval}
\end{table}
\cref{tab:wav2vec-eval} shows higher FGD and worse BeatAlign scores when using wav2vec, which also incurs an additional computation cost of computing audio embeddings.
In contrast, using Moshi~\cite{defossez2024moshi}'s internal text and speech token stream gives us better quantitative metrics and saves time for encoding and decoding speech.

\paragraph{Two-dimensional Transformer Design.}
We ablate our design choice of using a two-tier arrangement of temporal and kinematic transformers.
As discussed in~\cref{sec:method}, the disadvantage of using a single stream for both dimensions $T$ and $K$ is the scale-up in context-length of attention layers.
This manifests itself during training in terms of bad convergence, leading to overall worse performance in metrics.
\cref{tab:flatten} demonstrates that using a single transformer results in higher FID, worse BeatAlign scores and lower diversity.
Not to mention, the step times are almost doubled due to increased attention context.
\begin{table}[t]
\centering
\caption{\textbf{Comparison of Model Variants on Gesture Generation and Runtime.}}
\resizebox{\linewidth}{!}{%
\begin{tabular}{lcccc}
\toprule
 & $\text{FGD}\!\downarrow$
 & $\text{BeatAlign}\!\rightarrow$
 & $\text{L1-Div}\!\rightarrow$
 & $\text{Step Time (s)}\!\downarrow$ \\
\midrule
 GT                        
&       & 0.446 & 8.45 \\ 
\midrule
Single Transformer
& 1.256 & 0.731 & 5.48  & 0.096 \\

\textbf{Ours}
& 0.480 & 0.461 & 10.44  & 0.035 \\
\bottomrule
\end{tabular}
}

\label{tab:flatten}
\end{table}

\paragraph{Effect of additional losses.}
We ablate the contribution of auxiliary losses to our training by evaluating final models on the evaluation sets. 
Our base losses consist of $\mathcal{L}_{CE}$ and $\mathcal{L}_{\text{va}}$. 
We evaluate two different losses that are applied on estimated latents and ground-truth: (1) contrastive loss $\mathcal{L}_{\text{con}}$ and (2) MSE-loss.
\cref{tab:losses} shows that contrastive loss improves FGD from the base setup of cross-entropy loss, while applying direct MSE on estimated latents increases FGD.
\begin{table}[t]
\centering
\caption{\textbf{Quantitative Effect of Losses on Generation.}}
\resizebox{0.85\linewidth}{!}{%
\begin{tabular}{lccc}
\toprule
 & $\text{FGD}\!\downarrow$  
 & $\text{BeatAlign}\!\rightarrow$
 & $\text{L1-Div}\!\rightarrow$
 \\
 \midrule
GT                        
&       & 0.446 & 8.45 \\ 
\midrule
$\mathcal{L}_{CE}+\mathcal{L}_{\text{va}}$ 
& 0.499 & 0.450 & 10.25 \\

with MSE-loss
& 0.577 & 0.438 & 9.79 \\

with $\mathcal{L}_{\text{con}}$
& 0.480 & 0.461 & 10.44 \\
\bottomrule
\end{tabular}
}

\label{tab:losses}
\end{table}

\paragraph{Evaluation of Gesture Codec across $K$ levels.}
Since we divide our gesture token structure in $K$ levels to represent finer kinematic details, we evaluate how many of these levels are necessary for gesture tokenization.
\cref{tab:codebook} demonstrates the relation between the increasing number of levels and reconstruction quality.
We report Mean Per-Joint Position Error as a metric to evaluate the reconstruction quality for varying number of $K$ levels.
Lastly, we observe that generative FGD also follows a similar pattern as MPJPE.
\begin{table}[t]
\centering
\caption{\textbf{Effect of Number of Codebooks $K$ on Motion Reconstruction.} MPJPE is represented in meters.}
\resizebox{0.6\linewidth}{!}{%
\begin{tabular}{lcc}
\toprule
 & $\text{FGD}\!\downarrow$
 & $\text{MPJPE (m)}\!\downarrow$ \\
\midrule
$K{=}1$
& 0.55 
& 0.043 \\

$K{=}2$
& 0.42 
& 0.032 \\

$K{=}4$
& 0.135 
& 0.022 \\

$K{=}8$
& 0.059 
& 0.016 \\
\bottomrule
\end{tabular}
}

\label{tab:codebook}
\end{table}

\section{Limitations \& Future Work}
Our current framework models only the agent’s motion and does not incorporate the user’s body dynamics or full dyadic context, limiting its ability to handle interactive, multi-party gestures. Extending \model{} to perceive and respond to a partner’s gestures is an important direction for future work.
\section{Conclusion}
\label{sec:conclusion}
In this work, we present \model~-- an online, causal framework for generating expressive co-speech gestures and facial expressions synchronized with real-time dialogue. Through body-part–aware gesture codecs and a two-dimensional causal generator, our method models both temporal and kinematic motion structure at low latency. Contrastive objectives further enhance gesture diversity and expressiveness. Experiments across single- and multi-speaker settings show that \model{} produces natural, contextually aligned gestures and outperforms recent baselines. Our approach moves ECAs closer to truly interactive, human-like embodied communication.
\paragraph{Acknowledgments.}
This work was funded by the Deutsche Forschungsgemeinschaft (DFG, German Research Foundation) -- GRK 2853/1 “Neuroexplicit Models of Language, Vision, and Action” - project number 471607914.
We also thank Anton Zubekhin \& Andrea Boscolo Camiletto for their help with the demo.
{
    \small
    \bibliographystyle{ieeenat_fullname}
    \bibliography{main}
}
\clearpage
\setcounter{page}{1}
\maketitlesupplementary

\section{Online Generation Demo}
To demonstrate the real-time capabilities of our gesture generation framework, we build an interactive demo in which a user can converse naturally with an Embodied Conversational Agent (ECA).
We urge readers to watch the supplementary video, which highlights how our system supports online, continuous, and responsive gesture generation during live interaction.
\subsection{Architecture}
The primary goal of the demo is to showcase \model’s ability to generate gestures and speech in real time during a fully interactive conversation with the user.
Unlike traditional turn-based systems, our setup supports full-duplex interaction, allowing both the user and the ECA to speak, interrupt, and respond fluidly—mirroring natural human dyadic communication.
Achieving such responsiveness requires maintaining low latency while processing both the user's input and \model’s output continuously.
\par
\begin{figure}
    \centering
    \includegraphics[width=1.1\linewidth]{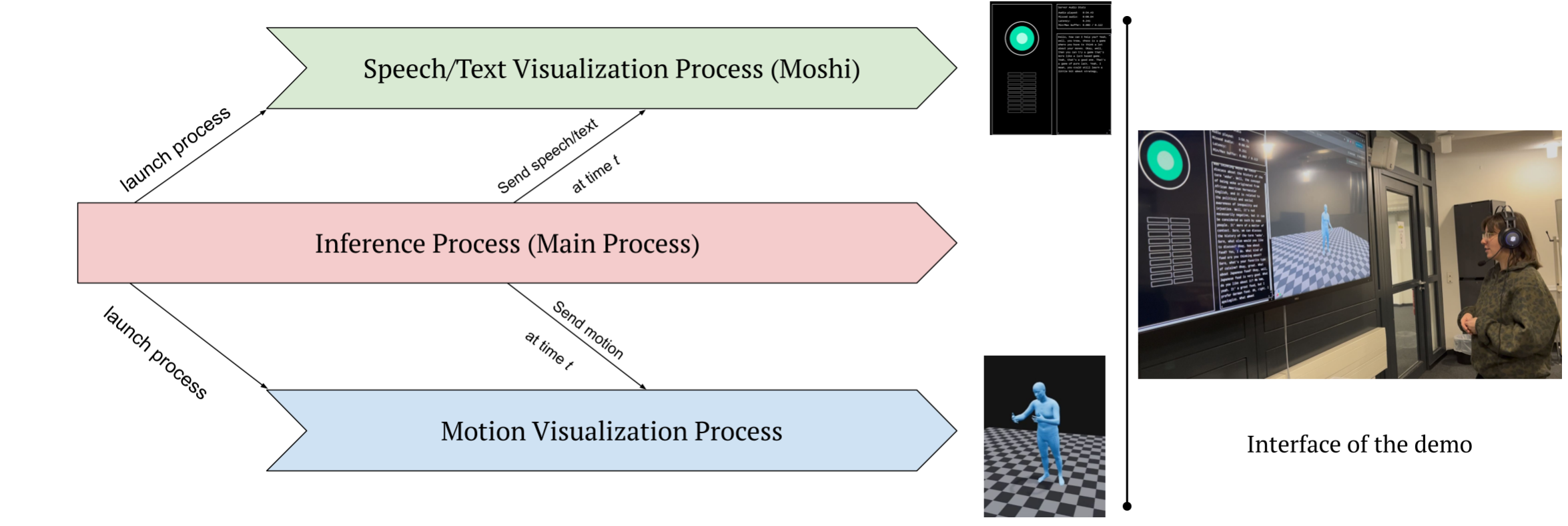}
    \caption{\textbf{System architecture of our real-time demo.} The main inference process runs Moshi and \model{} in a continuous loop, while two parallel processes handle speech/text visualization and motion rendering. Data is streamed between processes at each timestep via websockets to support low-latency, full-duplex interaction. Right: the user-facing interface of the demo.}
    \label{suppfig:arch}
\end{figure}

To this end, we implement the demo using three parallel processes, executed on a workstation equipped with an NVIDIA RTX 3090 GPU.
These processes run concurrently and communicate through lightweight websocket channels to ensure synchronized, low-overhead data exchange.
The three processes operate as follows:
\begin{itemize}
    \item \textbf{Inference Process (Main Process).}
This process runs the core inference loop for both Moshi~\cite{defossez2024moshi} and \model.
It handles real-time speech–text token streaming and generates gesture tokens frame by frame.
\item \textbf{Speech/Text Visualization Process.}
At every inference step, the raw audio waveform and the decoded text tokens are sent via websocket to this process.
It visualizes the user’s speech and the agent’s responses, allowing real-time inspection of the conversational flow.
\item \textbf{Motion Visualization Process.}
In parallel, the gesture generation module sends a time-aligned SMPL-X~\cite{SMPL-X:2019} mesh for each frame to a dedicated visualization process.
This process renders the full-body motion—including hands and facial expressions—on the user’s screen in real time.
 
\end{itemize}
Together, these components enable seamless, continuous interaction with the embodied agent, as illustrated in \cref{suppfig:arch}.
The system maintains low latency at each stage, enabling a fluid and immersive demonstration of real-time embodied dialogue.
\section{Additional Results on Embody3D~\cite{embody3d}}
Embody3D is a recently released dataset containing 59 hours of dyadic interaction recordings.
In this setup, two interlocutors face each other and communicate naturally, mirroring human–human conversational dynamics.
We evaluate on this dataset because our long-term goal is to develop fully interactive embodied agents capable of behaving like humans in real conversational settings.
\begin{table}[t]
\centering
\caption{\textbf{Quantitative evaluation on the Embody3D dataset.}}
\resizebox{0.6\linewidth}{!}{%
\begin{tabular}{lccc}
\toprule
 & $\text{FGD}\!\downarrow$
 & $\text{BeatAlign}\!\rightarrow$
 & $\text{L1-Div}\!\rightarrow$
 \\
\midrule
GT
& -- 
& 0.453 
& 5.97 
\\
\midrule
EMAGE~\cite{liu2024emage}
& 3.786 
& 0.022 
& 1.79 
\\
GestureLSM~\cite{liu2025gesturelsm}
& 3.744 
& 0.776 
& 13.75 
\\
\model
& \gold{1.642} 
& \gold{0.605} 
& \gold{10.18} 
\\
\bottomrule
\end{tabular}
}

\end{table}

\par
We finetune our multi-speaker BEAT2 models on Embody3D and report performance using FGD, BeatAlign, and L1 Divergence. Across all metrics, \model{} achieves the best quantitative results, showing lower gesture distribution divergence, improved alignment with speech prosody, and motion diversity closer to GT. This performance trend mirrors our findings on the BEAT2 multi-speaker evaluation, further demonstrating that our causal token-based framework generalizes well to new conversational settings. For fair comparison, we retrain the FGD network following EMAGE~\cite{liu2024emage} and recompute the mean velocity used in the BeatAlign metric. All evaluations are conducted using only the upper body and hands
\section{Analyzing Autoregressive Dependency in Kinematic Transformer.}
\begin{figure}   
\centering
\includegraphics[width=\columnwidth]{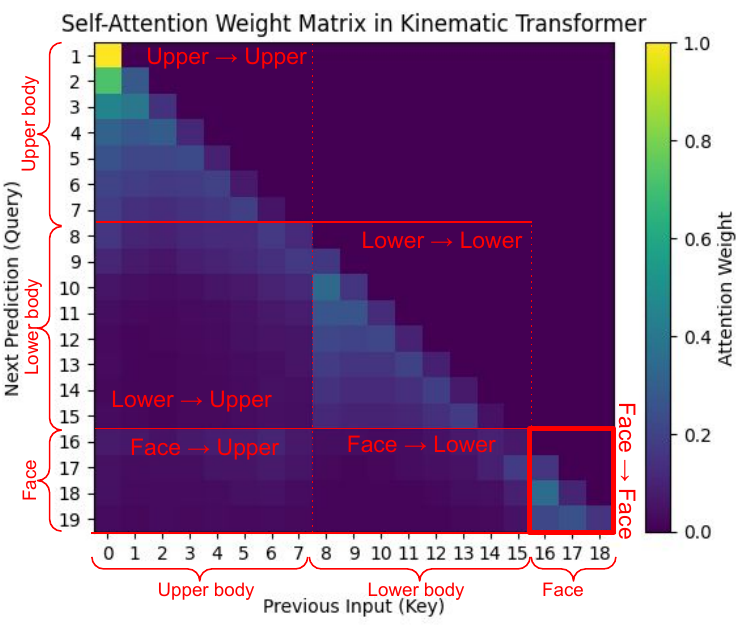}
\caption{\textbf{Kinematic Dependency Analysis.} Here, ``$\rightarrow$'' means ``attends to''.}
\label{fig:kinematicdep}
\end{figure}

Since we model body-part level details through an autoregressive transformer, this leads to a dependency in which body parts predicted later (lower body and face) depend on body parts predicted earlier (upper body).
Therefore, we analyze the effect of this ordering to examine whether it imposes a specific dependency chain.
We plot the causal self-attention between the face, upper-body, and lower-body tokens in~\cref{fig:kinematicdep}.
Even though face tokens are predicted after lower-body tokens, the attention weights show that the model implicitly learns to ignore lower-body tokens when predicting face tokens.
We observe that face self-attention is concentrated in the ``Face $\rightarrow$ Face'' block, as the face does not depend on other parts.
Lower-body tokens exhibit small attention to the upper body, since both are linked in terms of motion dynamics.
\section{On Causality-Quality Trade-off}
\label{sec:trade-off}
Recall that \model{} is designed to be an interactive, embodied conversational agent (ECA). 
This necessitates that the model is not only causal, but also real-time and our design choices have been profoundly dictated by these considerations.
Naturally, causality comes at the cost of quality.
However, this trade-off is not merely a consequence of having a limited context.
In this section, we highlight the underlying nuances associated with the design of causal synthesis for human gestures.
\par
\noindent\textbf{Where do Gestures originate?}
We submit that the premise of causal co-speech gesture synthesis is rather ill-posed, as it tacitly assumes that the agent (or humans) gesture on the basis of speech uttered within the \textit{past} context.
This assumption, however, is not true.
In reality, human speech and gestures are driven in parallel through a shared \textit{intent}~\cite{kendon2004gesture, deRuiter2000, Levelt1989, kita2003does}.
This is reflected in the observation that gestures can often be stroked even before the speech has been uttered~\cite{mcneill1992handmind, morrel1992word}. 
This is also observed in turn-taking between multiple interlocutors~\cite{suresh2025modeling}.
Likewise, it is also possible that gestures occur with a delay (for example, to reinforce or qualify the argument in the speech)~\cite{mcneill2005gesture}.
While the latter case can be, in principle, modeled by a causal model, the former is inherently challenging to achieve.
Consequently, we observe that causal modeling incentivizes more prominent beat gestures, as the temporal correlation between the speech and the gestures is easier to discover while training.
On the other hand, non-causal, full-context models, thrive in the luxury of future context availability and are able to model more nuanced and semantically meaningful gestures~\cite{zhang2024semanticgesture, mughal2025raggesture}.
\par
\noindent \textbf{Should we Model the Intent Behind the Gestures?}
We believe that a common modeling of the intent before generating the speech and gestures is a goal worth pursuing.
This could be achieved by first inferring the intent behind an LLM's output, and then generating the speech and gestures jointly based on the inferred intent.
However, this would be a suboptimal approach that breaks the constraints of causality, while also being slow.
For truly interactive ECAs, we either need a real-time LLM that is trained to generate the intent before the final output, or, we need an approach to disentangle the LLM's intent from the intermediate features of the model.
While fascinating directions for future research, both the potential solutions remain out of the scope of our current submission.
\section{Implementation Details of Gesture Codecs.}
We build streaming codecs for each body part using Residual VQ-VAE~\cite{zeghidour2021soundstream}.
These codecs consist of an encoder-decoder architecture, where the encoder downsamples the input motion sequence by a factor of 2 and the decoder upsamples it back for reconstruction.
Given an input of 250 frames during training, the encoder outputs 125 tokens for a 10-second sequence.
During training, the input sequence length is randomly sampled between 2 and 250 at every iteration.
\par
Upper and lower body codecs consist of 2 1d-convolutional layers and 8 transformer layers with 4 attention heads.
Face codec contains 2 1d-convolutional layers and 4 transformer layers with 2 attention heads.
Every codec is trained with a set of reconstruction and geometric losses along with commitment losses for each codebook.
We apply Geodesic Loss on rotation matrices and standard MSE losses on 6D, axis-angle and joint position representation of the motion.
Moreover, we also apply additional MSE losses to optimize
velocity/acceleration of motion~\cite{mughal2025raggesture}. Lastly, we apply loss on foot contact predictions during codec training to reduce
foot sliding~\cite{mughal2025raggesture, zhang2024roam}.
%
%
%

\section{Evaluation Metrics}
\paragraph{FGD.}
We adopt the Fréchet Gesture Distance (FGD), following Yoon~\etal~\cite{yoon2020trimodal}.
For evaluation, we use the gesture encoder released with BEAT2~\cite{liu2024emage} to extract gesture embeddings and compute FGD, without retraining the encoder.
\paragraph{Beat Alignment Score.}
Originally proposed to assess synchronization between music beats and dance motion~\cite{aichoreo}, the Beat Alignment Score has been adapted for gesture synthesis to measure how well gesture beat events align with audio beat events. It captures temporal correlation between gesture dynamics and speech prosody.
\paragraph{L1 Divergence.}
Also referred to as L1 variance, this metric computes the average L1 distance between each generated pose and the mean pose of the sequence. Lower values indicate motion collapse toward static poses, making it useful for detecting unexpressive or frozen gesture generation.
\paragraph{Facial-MSE.}
This metric, introduced by EMAGE~\cite{liu2024emage}, measures mean squared error between the ground truth facial expressions and predicted facial expressions. FLAME~\cite{FLAME:SiggraphAsia2017} is used as the representation to calculate this loss. 
\paragraph{Mean Per Joint Position Error (MPJPE).}
This is a standard metric for evaluating motion reconstruction and pose estimation. It measures the average Euclidean distance between predicted and ground-truth joint positions across all joints and frames. Formally, it is computed as the mean L2 distance in 3D space, providing a direct measure of pose accuracy.
\section{Details on User Study}
To evaluate perceptual quality, we conducted a user study with 53 participants.
Each participant was presented with 15 forced-choice questions randomly sampled from 45 questions. These questions display a side-by-side animation, comparing our method against state-of-the-art baselines and ground truth.
Each question displayed a side-by-side animation of our model and one of EMAGE~\cite{liu2024emage}, GestureLSM~\cite{liu2025gesturelsm}, or the ground truth.
For every pairwise comparison, participants answered two questions:
\begin{itemize}
    \item “Which gesture sequence looks more natural?”
    \item “Which appears better aligned with the spoken content?”
\end{itemize}
Across all comparisons, results were statistically significant with p-values $<0.001$, except for the appropriateness comparison against GestureLSM~\cite{liu2025gesturelsm}, which remained significant at $p<0.05$.
\section{Baseline Implementations}
For the single-speaker evaluation, we use the publicly released checkpoints provided by each baseline method.
For the multi-speaker evaluation, many baselines do not release multi-speaker models.
To ensure a fair comparison, we retrain EMAGE~\cite{liu2024emage} and MambaTalk~\cite{xu2024mambatalk} on the 23-speaker subset of BEAT2 (excluding \textit{carla} and \textit{itoi}), following the training configurations described in their respective papers.
\par
Beyond comparing to non-causal baselines, we also create causal variants of GestureLSM and MambaTalk to evaluate them under the same online, real-time constraints as our method.
In both cases, causality is enforced by applying a causal attention mask to all transformer layers during training.
This allows us to report quantitative comparisons against models operating under equivalent causal conditions.

\end{document}